\documentclass[letterpaper]{article} 
\usepackage[preprint]{aaai2027}  

\usepackage[hyphens]{url} 
\usepackage{graphicx} 
\usepackage{natbib}  
\usepackage{caption} 
\usepackage{booktabs}

\definecolor{mygray}{gray}{.9}
\usepackage{tabularx}

\usepackage{amsmath,amssymb}
\usepackage{amsthm}

\newcommand{\RPS}{RPS}
\newcommand{\RCS}{RCS}
\newcommand{\ESC}{ESC}
\newcommand{\droma}{d_{\mathrm{RoMa}}}
\usepackage{cases}

\usepackage{multirow}
\usepackage{xcolor}
\theoremstyle{plain} 
 
\usepackage[ruled]{algorithm2e}
\usepackage[export]{adjustbox}
\usepackage{threeparttable}
\usepackage{subcaption}
\usepackage{enumitem}
\usepackage{newfloat}
\usepackage{listings}
\usepackage{colortbl}
\usepackage{arydshln}
\usepackage{amsfonts}
\usepackage{pifont}
\usepackage{float}
\usepackage{bm}
\usepackage{mathrsfs}

\newcolumntype{x}[1]{>{\centering\arraybackslash}p{#1pt}}
\newcolumntype{I}{!{\vrule width 1pt}}

\usepackage{tcolorbox}
\tcbuselibrary{theorems}
\definecolor{lightgray}{gray}{.9}
\definecolor{deepgray}{gray}{.8}
\tcbset{highlight math/.append style={left=0mm,right=0mm,top=0mm,bottom=0mm, colframe=white}}
\definecolor{DarkBlue}{rgb}{0,0.08,0.45}
\definecolor{impgreen}{rgb}{0.05,0.50,0.20}
\newcommand{\dn}[1]{\,{\color{impgreen}\scriptsize$\downarrow${#1}\%}}
\newcommand{\up}[1]{\,{\color{impgreen}\scriptsize$+${#1}\%}}

\makeatletter
\newcommand{\thickhline}{%
    \noalign {\ifnum 0=`}\fi \hrule height 1pt
    \futurelet \reserved@a \@xhline
}



\theoremstyle{plain}

\theoremstyle{definition}

\theoremstyle{remark}

\usepackage{xspace}
\makeatletter
\DeclareRobustCommand\onedot{\futurelet\@let@token\@onedot}
\def\@onedot{\ifx\@let@token.\else.\null\fi\xspace}

\title{WorldCycle: Self-Verifiable Reinforcement Learning for Long-Horizon Video World Models}
\author{
Bohai Gu\textsuperscript{1,\equalcontrib},
Yueyang Yuan\textsuperscript{2,\equalcontrib},
Taiyi Wu\textsuperscript{3},
Dazhao Du\textsuperscript{1},
Jian Liu\textsuperscript{1},
Xiaoyi Pang\textsuperscript{1},
Jie Zhang\textsuperscript{1},
Xiaocheng Lu\textsuperscript{1},
Haobin Zhong\textsuperscript{3},
Xiaotong Zhao\textsuperscript{3},
Alan Zhao\textsuperscript{3},
Song Guo\textsuperscript{1,\thanks{Corresponding author}}
}

\affiliations{
\textsuperscript{1}The Hong Kong University of Science and Technology
\textsuperscript{2}Wuhan University \\
\textsuperscript{3}AI Technology Center, Tencent Video, Tencent}

\begin{document}

\maketitle

\begin{abstract}

Interactive video world models are essential for long-horizon planning and exploration, yet they suffer from compounding errors. Post-training methods such as reinforcement learning (RL) can improve these models, but they hit a verification bottleneck: for arbitrary action sequences, no ground-truth future state exists to measure long-term drift. Our key insight is that reversible action cycles make this verification possible: a sequence composed with its inverse must analytically return to the initial state, yielding annotation-free supervision on long-horizon correctness. Building on this, we introduce \textit{\textbf{WorldCycle}}, a self-verifiable RL framework that constructs closed action cycles and their repeated executions from ordinary action sequences, and optimizes two complementary rewards: a \textbf{spatial closure reward} enforcing symmetry between mirrored forward and reverse segments, and a \textbf{temporal consistency reward} aligning states across repeated cycle executions. 
These rewards force the model to learn actions as consistent state operators rather than memorized temporal patterns, and extend naturally to out-of-distribution composite action cycles that the base model handles poorly. We further release \textit{\textbf{CycleBench}}, a diagnostic benchmark for state-returning ability under complex action structures. 
WorldCycle reduces state returning drift by up to 44\% and lifts composite-action accuracy nearly \textbf{4$\times$} over the base model, providing a vital foundation for physically grounded world models.

\end{abstract}

\begin{figure*}[t]
\centering
\includegraphics[width=\linewidth]{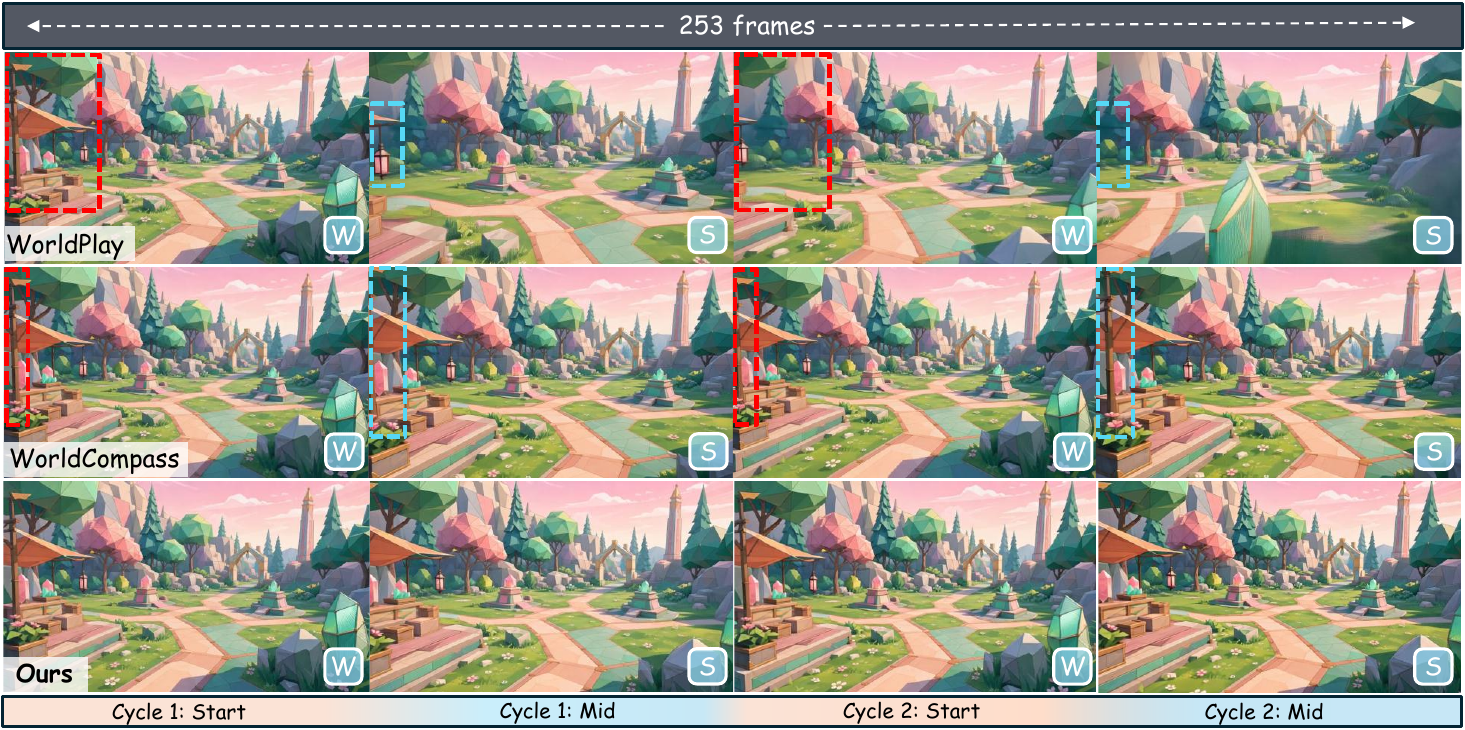}
\vspace{-20pt}
\caption{
  \textbf{State-returning failure in interactive video world models.}
  Even state-of-the-art models fail a minimal sanity check: an inverse
  action sequence such as forward-then-backward does not recover the
  starting view (\textcolor[rgb]{1,0.3,0.4}{\textbf{spatial closure failure}}), and executing the same action from what should be the same state at different rollout steps
  produces inconsistent displacements (\textcolor[rgb]{0.53,0.80,0.92}{\textbf{temporal consistency failure}}). Both failures accumulate
  as autoregressive rollouts extend, yet neither is captured by
  short-horizon rewards. WorldCycle turns closed action cycles into
  annotation-free supervision on long-horizon correctness.
}
\label{fig:teaser}
\vspace{-20pt}
\end{figure*}

\section{Introduction}

Interactive video world models (IWMs) generate future visual observations conditioned on an initial observation and a stream of user or agent actions, turning generative video models into interactive simulators~\cite{yang2024unisim,wu2024ivideogpt,zhu2025irasim,xiang2024pandora}. Such models support embodied interaction, robot learning, visual planning, and game-like simulation. Related action-conditioned video world models have also been developed for autonomous driving~\cite{hu2023gaia1,wang2024drivewm,gao2024vista,zhang2025epona}. These models commonly employ an autoregressive generation paradigm: each newly generated observation becomes the context for predicting the next one. Consequently, small single-step errors can propagate and compound as the rollout extends, gradually eroding the physical and geometric consistency required by downstream long-horizon tasks~\cite{bengio2015scheduled,janner2019mbpo,huang2026live,su2026cycleworld}.

Post-training methods based on reinforcement learning (RL) have recently emerged as a promising approach for improving video world models~\cite{wu2025rlvrworld,ye2025rlir}. Yet existing approaches predominantly optimize short-horizon visual quality or per-step action alignment. \textbf{The fundamental obstacle to long-horizon improvement is the absence of effective supervision.} For an arbitrary action sequence, no ground-truth future state is available to measure accumulated error. The most recent work, WorldCompass~\cite{wang2026worldcompass}, attempts to address this by rewarding clip-level action following, but such objectives primarily improve local action accuracy and remain insufficient for correcting fine-grained spatial drift over long horizons.

We identify a new angle on this problem by focusing on a specific and ubiquitous class of dynamics: reversible trajectories such as camera ego-motion. For a reversible action cycle, physical law dictates that the final state must be identical to the initial state, providing an absolutely precise, annotation-free reference. Yet as shown in Figure~\ref{fig:teaser}, both WorldPlay and WorldCompass~\cite{sun2025worldplay,wang2026worldcompass} fail this minimal check: an inverse action sequence fails to recover the starting view (\textbf{spatial closure failure}), and the same action produces different displacements at different rollout positions (\textbf{temporal consistency failure}). These failures compound with horizon length and are invisible to any short-horizon reward. Crucially, the same physical prior that exposes the failure also resolves the supervision bottleneck: the \textbf{cycle-return constraint} provides an exact, label-free target for the long-horizon trajectory. Starting from this insight, we identify two remaining challenges: \emph{(i) reward sparsity}, where an endpoint-only cycle signal provides too little gradient to localise intermediate errors; and \emph{(ii) temporal drift}, where AR accumulation makes the same action behave differently at different rollout depths. The harder second level concerns composite actions such as moving forward while turning, which must also return after their inverse programs. Pre-trained baselines such as WorldPlay exhibit a $5{\times}$ accuracy collapse on composite-action versus simple-action settings of identical rollout length (Table~\ref{tab:main_results}), indicating that composite motions are out-of-domain: pretraining data rarely provides GT video demonstrations of composite trajectories.

We propose \textit{\textbf{WorldCycle}}, a Self-Verifiable Reinforcement Learning framework specifically designed to tackle state-returning failure in long-horizon video world models under \textbf{reversible action trajectories}. The guiding principle is to \textbf{turn closed action programs into dense supervision}. We exploit the mirror structure of an inverse action program: every partial forward trajectory has a corresponding reverse state that should coincide with it. We construct mirrored frame pairs at every intermediate depth of a cycle and compare them with dense visual evidence, yielding \textbf{a spatial closure reward} that converts each intermediate chunk into an independently verifiable closure check, matching the chunk-wise structure of AR post-training and removing the need for a single sparse endpoint signal. We further repeat cycles and compare co-indexed frames across cycles, producing a \textbf{temporal state consistency reward} that penalises drift of identical actions over time, directly targeting the temporal drift obstacle.
These two rewards force the model to learn actions as consistent state operators rather than memorized temporal patterns.
The same cycle-based objective also applies to composite actions. Although the base model may not reliably execute such action combinations, WorldCycle can directly optimize them through cycle closure rewards, \textbf{without requiring ground-truth video supervision}.

We further introduce \textit{\textbf{CycleBench}}, a benchmark suite covering inverse, repeated, and composite cycles, enabling video IWMs to be evaluated as simulators rather than only as short-horizon generators. Because no existing benchmark diagnoses state-returning consistency, existing evaluations ask whether the model follows actions correctly, not whether its induced state-transition system is cycle-consistent~\cite{ying2026wbench,ye2026mind}. CycleBench fills this gap with four task types and four scenario settings spanning short-, mid-, and long-horizon lengths as well as composite-action conditions. Our contributions are threefold:

\begin{itemize}[leftmargin=1.4em,itemsep=1pt,topsep=1pt]
\item[\ding{182}] \textbf{A new long-horizon supervision perspective.} We reveal state-returning failures in long-horizon video IWMs and introduce reversible action cycles as a self-verifiable supervision signal, requiring no ground-truth trajectories.

\item[\ding{183}] \textbf{\textit{WorldCycle}: a self-verifiable RL framework for reversible dynamics.} We introduce two complementary trajectory-level rewards, spatial cycle closure and temporal state consistency, that overcome autoregressive drift on in-domain actions and generalize to out-of-distribution composite action cycles.

\item[\ding{184}] \textbf{\textit{CycleBench }and empirical gains.} We establish the first benchmark for state-returning consistency across reversible, repeated, and composite actions. WorldCycle reduces state returning drift by up to 44\% and improves composite-action accuracy nearly 4$\times$ over the baseline.
\end{itemize}

\vspace{-5pt}
\section{Related Work}

\textbf{Video World Models.}
Classical world models learn predictive environment dynamics for control and
planning, typically through compact internal or latent representations
\cite{ha2018world,hafner2019planet,schrittwieser2020muzero,
hafner2025mastering}.
Recent video world models instead represent predicted environment evolution
directly in pixel space, building on advances in video generation and visual
world modeling~\cite{ho2022video,kondratyuk2024videopoet,brooks2024sora,alonso2024diamond}. Interactive systems such as
Genie, GameNGen, Oasis, WorldPlay, and HY-World condition generation on user
actions for longer-horizon or real-time
interaction~\cite{bruce2024genie,valevski2024gamengen,decart2024oasis,sun2025worldplay,hyworld2026,parkerholder2024genie2,deepmind2025genie3,zhang2025matrixgame,gao2026lingbotworld2}.
These models pursue fidelity along a single forward direction: each step
should look right and follow the current action. Long-range coherence, when
addressed, is treated as an appearance- or geometry-level
property~\cite{chen2025vrag,wu2025spatialmemory,yan2023teco,po2025longcontext}. None of these objectives
can ask whether the rollout has reached the state the actions demand, because
along an arbitrary open trajectory no reference state exists. Reversible and
closed action trajectories are the exception: their composed transformation
must be the identity, fixing the correct outcome analytically and without
annotation. WorldCycle turns this structure into a trajectory-level training
signal.

\textbf{Reinforcement Learning for World Model Post-training.}
RL-based post-training has produced strong gains in language models, from
RLHF to GRPO~\cite{ouyang2022training,shao2024deepseekmath,guo2025deepseekr1}, and the paradigm has
been extended to diffusion models via DDPO, AlignProp, DiffusionDPO, Flow-GRPO, DanceGRPO,
and
DiffusionNFT~\cite{black2024ddpo,prabhudesai2024alignprop,wallace2024diffusiondpo,
liu2025flowgrpo,xue2025dancegrpo,zheng2025diffusionnft}.
For world models specifically, RLVR-World and RLIR apply RL post-training
with verifiable or inverse-dynamics
rewards~\cite{wu2025rlvrworld,ye2025rlir}, and WorldCompass scores each
generated clip against an inverse-pose action-following reward and a
visual-quality preference model~\cite{wang2026worldcompass}. All existing
methods supervise individual clips or short segments against an external
scorer, placing no constraint on the trajectory as a whole. The correct
long-horizon state is unknown for arbitrary action sequences, so
trajectory-level error cannot be penalized directly. WorldCycle resolves
this by constructing rewards whose target is fixed by the algebraic identity
of reversible cycles, yielding a label-free objective that
requires no external annotation.

\vspace{-5pt}
\section{Method}

\begin{figure*}[t]
\centering
\includegraphics[width=\linewidth]{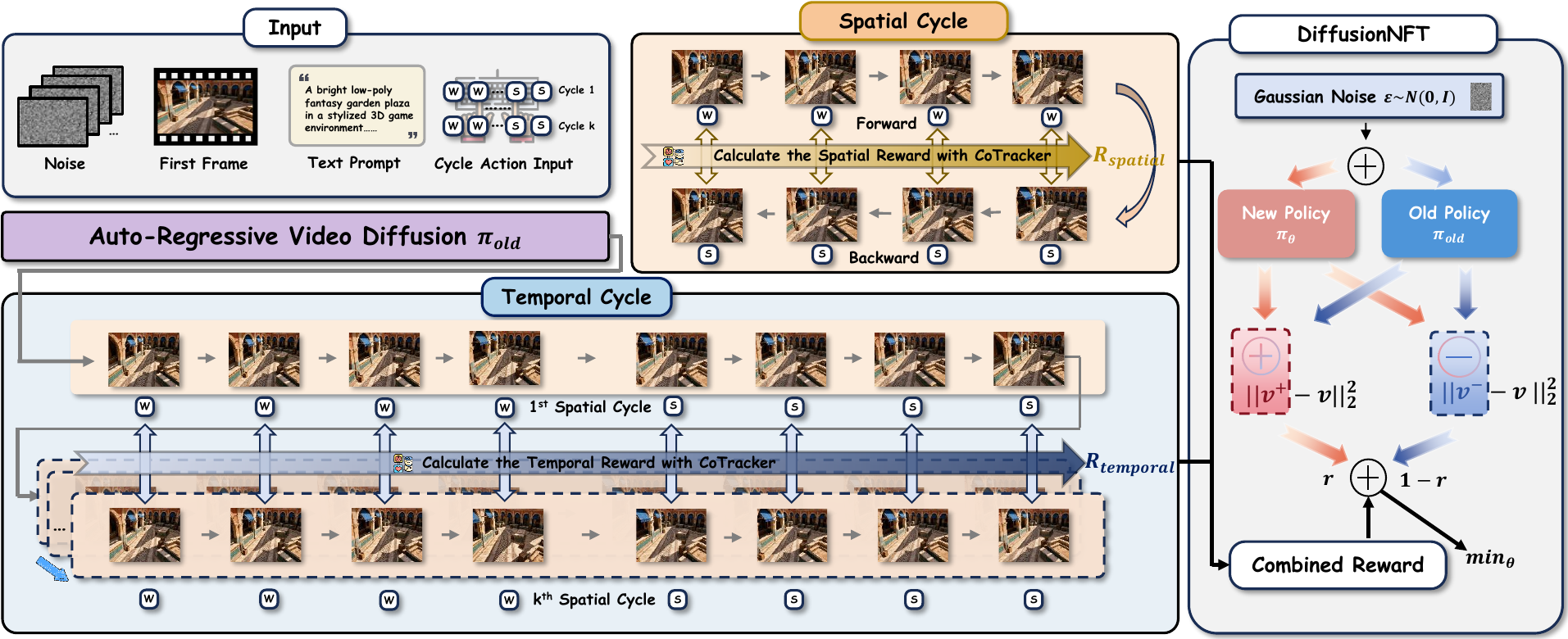}
\vspace{-20pt}
\caption{
  \textbf{Overview of WorldCycle.} From ordinary action sequences we
  construct closed reversible cycles and their repeated executions. The
  \emph{spatial closure reward} compares every mirrored forward-reverse
  frame pair within a single cycle, providing dense supervision on where
  the trajectory begins to drift. The \emph{temporal consistency reward}
  compares phase-aligned frames of the same action across repeated
  cycles, penalizing drift of identical actions over time. Both rewards
  are trajectory-level, annotation-free, and jointly optimized with a
  DiffusionNFT-style objective.
}
\label{fig:method}
\vspace{-20pt}
\end{figure*}

\subsection{Preliminaries}

\noindent\textbf{Interactive Video World Models.}
We formulate an interactive video world model as a learned state-transition
system $\mathcal{M}_{\theta}=(\mathcal{S},\mathcal{A},f_{\theta})$, where
$f_{\theta}:\mathcal{S}\times\mathcal{A}\rightarrow\mathcal{S}$ maps a state
and an action to the next state, $s_{t+1}=f_{\theta}(s_t,a_t)$. For a
reversible action $a$, let $T_a:\mathcal{S}\rightarrow\mathcal{S}$ denote its
true state transformation and $\hat{T}_a$ the transformation realized by the
learned model. Given a trajectory $\gamma=(a_1,\ldots,a_T)$, we define
$T_{\gamma}=T_{a_T}\circ\cdots\circ T_{a_1}$ and define
$\hat{T}_{\gamma}$ analogously. Even a well-trained model realizes each
action only up to a residual transformation,
$\hat{T}_a=T_a\circ E_a$ with $E_a\approx I$. Consequently,
\vspace{-5pt}
\begin{equation}
\begin{aligned}
\hat{T}_{\gamma}
&=
\hat{T}_{a_T}\circ\cdots\circ\hat{T}_{a_1} \\
&=
(T_{a_T}\circ E_{a_T})\circ\cdots\circ
(T_{a_1}\circ E_{a_1})
\neq T_{\gamma}.
\end{aligned}
\end{equation}

Because autoregressive predictions become the context for subsequent
generation, these residual transformations generally do not cancel and
instead compound with the rollout horizon. Directly measuring the resulting
deviation would require access to the correct transformation $T_{\gamma}$,
which is unknown for an arbitrary action sequence at training time. We refer
to this as the \emph{verification bottleneck}: long-horizon transition error
exists, but no reference state is available against which it can be directly
scored, forcing existing post-training objectives to rely on locally
verifiable proxies.

\noindent\textbf{Action-induced Transformation Group.}
The verification bottleneck admits a structural exception. We consider the
subset of reversible interactions whose induced transformations form a group
$\mathcal{G}=(\mathcal{T},\circ)$. Every action $a$ admits an inverse
$a^{-1}$, while the inverse trajectory $\gamma^{-1}$ cancels the complete
transformation induced by $\gamma$:
\vspace{-3pt}
\begin{equation}
T_a\circ T_{a^{-1}}=I,
\qquad
T_{\gamma^{-1}\circ\gamma}=I.
\end{equation}

The identity target is determined entirely by the algebraic structure of the
actions and is therefore independent of scene content, the initial state, and
intermediate observations. Closed reversible trajectories consequently turn
an otherwise unknown long-horizon target into an analytically known one,
providing a content-independent and annotation-free verifier for accumulated
transition error.

\vspace{-5pt}
\subsection{Self-Verifiable Reinforcement Learning}

WorldCycle converts this analytical identity constraint into a
trajectory-level training signal. For a closed trajectory $\gamma$,
recurrently executing the learned transition function, denoted by
$f_{\theta}(s,\gamma)$, should recover the state from which the trajectory
started. We define \emph{state returning} through the condition
\vspace{-3pt}
\begin{equation}
T_{\gamma}=I
\quad\Longrightarrow\quad
f_{\theta}(s,\gamma)=s,
\qquad
\forall s\in\mathcal{S}.
\end{equation}

This condition is stronger than local action following. A model may respond
correctly to every individual action while its composed transformation
$\hat{T}_{\gamma}$ still deviates from identity because of accumulated
residuals. We therefore construct two complementary objectives. The spatial
objective densely verifies mirrored states within a reversible cycle,
whereas the temporal objective verifies whether corresponding cycle phases
remain stable across repeated executions.

\noindent\textbf{State Discrepancy Measurement.}

Because the underlying physical state $s$ is inaccessible, we leverage dense point tracking within the image space as an observable proxy to measure state consistency.
Given two frames $o_i$ and
$o_j$, we track $N$ valid correspondence points and denote their image
coordinates by $p_n^i, p_n^j$. The frame-pair discrepancy is the
size-normalized mean displacement
\vspace{-5pt}
\begin{equation}
d(o_i,o_j)
=
\frac{1}{N(H+W)}
\sum_{n=1}^{N}
\left\lVert p_n^i-p_n^j\right\rVert_2,
\end{equation}
where $H,W$ are the image dimensions. Invalid or low-confidence matches are
excluded.

\noindent\textbf{Spatial Closure Reward.}
We construct a symmetric cycle consisting of a forward action sequence
followed by its exact inverse,
$\gamma_c=(a_1,\ldots,a_m,a_m^{-1},\ldots,a_1^{-1})$, for which
$T_{\gamma_c}=I$. The state reached after the first $i$ forward actions
should match the state reached at the mirrored phase $2m-i$ of the reverse
trajectory. Rather than supervising only the final endpoint, we use every
mirrored pair to provide dense evidence throughout the rollout:
\begin{equation}
R_{\mathrm{spatial}}
=
\frac{1}{m+1}
\sum_{i=0}^{m}
\exp\!\left[
-\alpha d(o_i,o_{2m-i})
\right].
\end{equation}
\vspace{-10pt}

Here $o_i$ denotes the generated observation at phase $i$, and $d(\cdot,\cdot)$ measures the discrepancy between two observations that should correspond to the same underlying state. This mirrored supervision localizes where a reversible trajectory begins to drift, avoiding the sparse credit assignment produced by an endpoint-only reward.

\noindent\textbf{Temporal State Consistency Reward.}
Spatial closure verifies state agreement within a single cycle, but does not
guarantee that the recovered state behaves consistently when the cycle is
executed again. A valid state-transition system should satisfy
$s_t=s_{t'}\Rightarrow
f_{\theta}(s_t,a)=f_{\theta}(s_{t'},a)$. Identical states under identical actions must induce identical transitions.
To enforce this, we repeat the cyclic action program for $K$ cycles. Let $o_k^i$ denote the observation at step $i$ of cycle $k$ (where each cycle has length $L$). We anchor all subsequent cycles to the first cycle and define the reward as:
\vspace{-5pt}
\begin{equation}
R_{\mathrm{temporal}}
=
\frac{1}{(K-1)(L+1)}
\sum_{k=2}^{K}
\sum_{i=0}^{L}
\exp\!\left[
-\alpha d(o_k^i,o_1^i)
\right].
\end{equation}

Using the first cycle as a shared anchor avoids allowing the reference state to drift through adjacent-cycle comparisons. The resulting reward verifies not only whether each cycle  closes, but also whether repeated returns produce consistent subsequent evolution over an extended horizon.

\vspace{-5pt}
\subsection{Cycle-Consistent RL Training }

Trajectory-level cycle rewards require long rollouts, which are computationally expensive and challenging for standard RL algorithms to optimize. To make training practical, we introduce a structured curriculum to prevent learning shortcuts, alongside an efficient diffusion RL objective.

\noindent\textbf{Warm-up-and-Combine Schedule.}
Temporal consistency is unreliable if individual cycles do not close: phase-aligned comparisons across cycles are uninterpretable when the cycle has not been grounded.
We therefore begin with a spatial-only warm-up ($t<t_{\mathrm w}$) before
activating the temporal reward.
\begin{equation}
R_{\mathrm{cycle}}^{(i)}(t)
=
\lambda_s R_{\mathrm{spatial}}^{(i)}
+
\mathbf{1}[t\!\geq\! t_{\mathrm w}]\,
\lambda_t R_{\mathrm{temporal}}^{(i)}.
\end{equation}
After $t_{\mathrm w}$, both rewards remain jointly active.
A sequential strategy that switches to temporal-only risks catastrophic forgetting of spatial closure~\cite{kirkpatrick2017overcoming,kaplanis2019policy,yu2020gradient}.
An alternating strategy avoids forgetting but introduces gradient conflict: each objective switch partially reverses the previous optimization direction.
Joint optimization resolves both issues simultaneously: spatial closure provides the geometric grounding on which temporal consistency depends, and the two gradients reinforce rather than compete with each other.

\begin{figure*}[t]
\centering
\includegraphics[width=\linewidth]{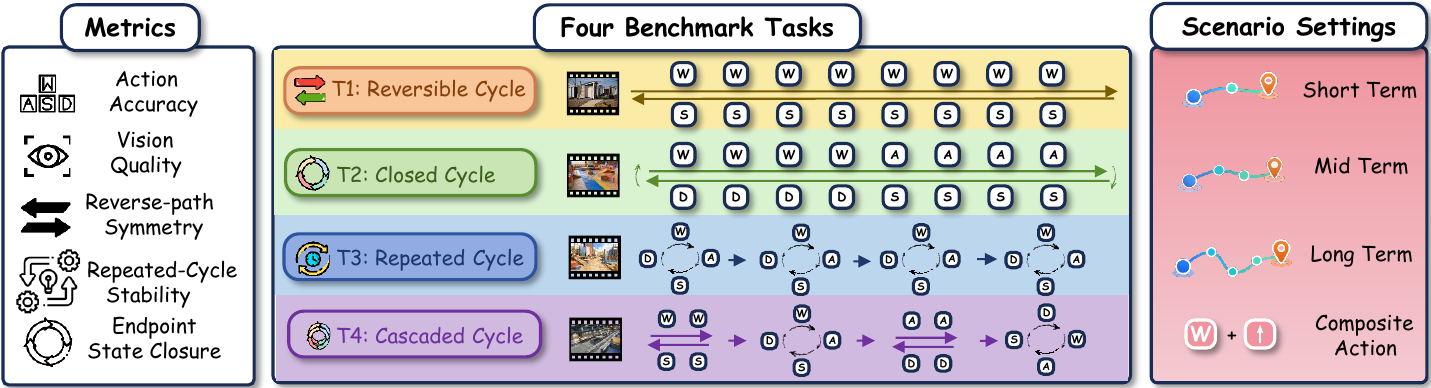}
\vspace{-20pt}
\caption{
  \textbf{Overview of CycleBench.} Four complementary tasks probe different
  aspects of accumulated transition error: T1 measures where error appears
  along a reversible path; T2 measures the residual after a non-symmetric
  closed trajectory; T3 tracks error growth across repeated cycles; T4
  measures error propagation across structurally different cycles. All
  tasks are further evaluated under three horizon settings (short/mid/long)
  and one out-of-domain composite-action setting.
}
\label{fig:benchmark}
\vspace{-20pt}
\end{figure*}

\noindent\textbf{Multi-Scale Cycle Sampling.}
A fixed cycle length may introduce a temporal-position shortcut, where the
model memorizes when to reverse rather than learning the inverse relation
between actions. We therefore enforce cycle closure at multiple temporal
scales. Given a rollout horizon $H=n\cdot2^Q$, where $n$ is the residual
factor and $2^Q$ denotes the largest power-of-two factor of $H$, we construct
hierarchical cycles by partitioning the rollout into $2^k$ ($k \in \{0,\ldots,Q-1\}$) cycles of length
$H/2^k$, each consisting of a forward segment followed by its inverse.
We randomly sample one hierarchy level $k$ during training, thus preventing reward hacking.

\noindent\textbf{Reward Composition.}
The cycle reward is the trajectory-level backbone of our objective, but on
its own it under-constrains two ingredients that a video world model must
still get right: local action fidelity and per-frame visual quality.
To ensure the cycle reward is not satisfied through visually degenerate shortcuts, we compose the final training reward from three complementary
signals:
\begin{equation}
R^{(i)}(t)
=
R_{\mathrm{cycle}}^{(i)}(t)
+
\lambda_a R_{\mathrm{act}}^{(i)}
+
\lambda_v R_{\mathrm{HPS}}^{(i)},
\end{equation}
where $R_{\mathrm{act}}$ is an action-following score~\cite{wang2026worldcompass}
and $R_{\mathrm{HPS}}$ is a visual-quality score~\cite{ma2025hpsv3}.

\noindent\textbf{Trajectory-Level Optimization with Efficient Rollouts.}
Cycle-consistency is a global property: rewards are computed over
the complete rollout before optimization. The reward design is
optimizer-agnostic and can be integrated into any diffusion RL framework
supporting autoregressive rollout. However, policy-gradient methods such as
DanceGRPO~\cite{xue2025dancegrpo} require timestep-wise log-probability
tracking and often reduce rollout diversity under long-horizon optimization
~\cite{wang2026worldcompass}. We therefore adopt a DiffusionNFT-style
negative-aware objective~\cite{zheng2025diffusionnft}, which performs
supervised regression toward reward-weighted positive and negative velocity
directions:
\begin{equation}
\mathcal{L}_{\mathrm{RL}}^{(i)}
=
r^{(i)}\|v_\theta^{+}-v^{(i)}\|_2^2
+
(1-r^{(i)})\|v_\theta^{-}-v^{(i)}\|_2^2 ,
\end{equation}
where $r^{(i)}=\operatorname{Norm}(R^{(i)})$ is the normalized cycle reward.
Following existing work~\cite{wang2026worldcompass}, we backpropagate only through the trailing chunk while keeping previous chunks as frozen context to reduce memory cost for long rollouts, preserving trajectory-level supervision with a per-step cost comparable to
single-clip training.

\noindent\textbf{Analysis on Composite Action Generalization.}
Composite actions lie outside the
base model's pretraining distribution.
WorldCycle addresses this directly: a composite action sequence composed with its
inverse still forms a closed cycle, so the cycle-closure reward provides
annotation-free supervision on composite trajectories without requiring any
ground-truth video.
The mechanism that makes this effective is transition residual reduction.
Consider a composite trajectory
$\gamma_{\mathrm{comp}}=(c_1,\ldots,c_T)$, where each $c_t$ may be a primitive
action or a combination of multiple motion components. Its learned rollout can
be written as
\begin{equation}
\vspace{-5pt}
\hat{T}_{\gamma_{\mathrm{comp}}}
=
T_{\gamma_{\mathrm{comp}}}
\circ
\mathcal{E}_{\gamma_{\mathrm{comp}}},
\end{equation}
where $\mathcal{E}_{\gamma_{\mathrm{comp}}}$ denotes the accumulated transition
residual along the trajectory. For a nominally closed composite trajectory,
$T_{\gamma_{\mathrm{comp}}}=I$, and the final drift is therefore determined by
the residual accumulation. By optimizing cycle closure, WorldCycle reduces
local transition errors and their propagation across action sequences, thereby
encouraging the model to recover the underlying action composition algebra
rather than memorizing specific trajectories. Consequently, the learned
dynamics generalize to unseen combinations of translations, rotations, and
their compositions.

\begin{table*}[t]
\centering
\footnotesize
\caption{
\textbf{CycleBench results} (weighted average over four benchmark tasks). Model size and venue are indicated after each baseline. WorldCompass and WorldCycle are both post-trained from WorldPlay; reported {\color{impgreen}gains} are relative to WorldCompass.
}
\vspace{-10pt}
\label{tab:main_results}
\resizebox{\linewidth}{!}{%
\renewcommand{\arraystretch}{1.25}
\setlength{\minrowclearance}{2pt}
\setlength{\tabcolsep}{4pt}
\begin{tabular}{l ccccc I ccccc}
\toprule
& \multicolumn{5}{c I}{\textbf{Short-term (125 frames)}}
& \multicolumn{5}{c}{\textbf{Mid-term (253 frames)}} \\
\cmidrule(lr){2-6}\cmidrule(lr){7-11}
\rowcolor{mygray}\textbf{Method}
  & ESC$\downarrow$ & RPS$\downarrow$ & RCS$\downarrow$ & Acc$\uparrow$ & Qual$\uparrow$
  & ESC$\downarrow$ & RPS$\downarrow$ & RCS$\downarrow$ & Acc$\uparrow$ & Qual$\uparrow$ \\
\midrule
Lingbot World v2 \textit{\scriptsize(14B)}
  & 0.198 & 0.143 & 0.085 & 0.464 & \textbf{12.11}
  & 0.227 & 0.233 & 0.062 & 0.480 & 10.58 \\
\rowcolor{gray!8}WorldPlay \textit{\scriptsize(8B, ICML'26)}
  & 0.076 & 0.057 & 0.066 & 0.635 & 9.66
  & 0.092 & 0.065 & 0.053 & 0.695 & 9.00 \\
WorldCompass \textit{\scriptsize(ICML'26)}
  & 0.038 & 0.034 & 0.026 & 0.829 & 11.06
  & 0.048 & 0.039 & 0.025 & 0.868 & 10.67 \\
\hdashline
\rowcolor[HTML]{D7F6FF}\textbf{WorldCycle}
  & \textbf{0.026}\dn{32} & \textbf{0.019}\dn{44} & \textbf{0.020}\dn{23} & \textbf{0.833}\up{0.5} & 11.24\up{2}
  & \textbf{0.038}\dn{21} & \textbf{0.028}\dn{28} & \textbf{0.018}\dn{28} & \textbf{0.878}\up{1} & \textbf{10.81}\up{1} \\
\specialrule{1.2pt}{2pt}{2pt}
& \multicolumn{5}{c I}{\textbf{Composite-action (125 frames)}}
& \multicolumn{5}{c}{\textbf{Long-term (381 frames)}} \\
\cmidrule(lr){2-6}\cmidrule(lr){7-11}
\rowcolor{mygray}\textbf{Method}
  & ESC$\downarrow$ & RPS$\downarrow$ & RCS$\downarrow$ & Acc$\uparrow$ & Qual$\uparrow$
  & ESC$\downarrow$ & RPS$\downarrow$ & RCS$\downarrow$ & Acc$\uparrow$ & Qual$\uparrow$ \\
\midrule
Lingbot World v2 \textit{\scriptsize(14B)}
  & 0.204 & 0.143 & 0.102 & 0.414 & \textbf{10.81}
  & 0.221 & 0.186 & 0.123 & 0.033 & \textbf{11.73} \\
\rowcolor{gray!8}WorldPlay \textit{\scriptsize(8B, ICML'26)}
  & 0.119 & 0.088 & 0.057 & 0.136 & 8.98
  & 0.302 & 0.202 & 0.173 & 0.073 & 4.71 \\
WorldCompass \textit{\scriptsize(ICML'26)}
  & 0.073 & 0.059 & 0.040 & 0.499 & 10.12
  & 0.187 & 0.104 & 0.074 & 0.084 & 9.81 \\
\hdashline
\rowcolor[HTML]{D7F6FF}\textbf{WorldCycle}
  & \textbf{0.057}\dn{22} & \textbf{0.042}\dn{29} & \textbf{0.026}\dn{35} & \textbf{0.553}\up{11} & 10.42\up{3}
  & \textbf{0.163}\dn{13} & \textbf{0.087}\dn{16} & \textbf{0.049}\dn{34} & \textbf{0.095}\up{13} & 10.31\up{5} \\
\bottomrule
\end{tabular}%
}
\vspace{-15pt}
\end{table*}

\vspace{-5pt}
\section{CycleBench}

CycleBench measures \textbf{how transition errors accumulate over extended
autoregressive rollouts}. For an arbitrary open trajectory, the correct
future state is unavailable, so long-horizon error cannot be scored
directly. CycleBench sidesteps this by constructing reversible and closed
action trajectories whose composed transformation is analytically the
identity: any discrepancy between rollout states that should coincide is
an observable estimate of accumulated error. The benchmark consists of 47
action trajectories evaluated from 380 initial frames, organized as
\textbf{four tasks} $\times$ \textbf{four settings}, and provides
correspondence-based metrics that localize where errors emerge and how
they propagate.

\noindent\textbf{Benchmark Tasks.} Each task uses analytically known frame correspondences within closed
trajectories to probe a distinct aspect of error accumulation.
\textbf{T1 (Reversible-Cycle)} executes a forward action sequence followed
by its exact inverse and evaluates whether mirrored forward-reverse frames
coincide, providing a dense path-wise probe.
\textbf{T2 (Closed-Cycle)} executes a non-retracing closed path (e.g., a
rectangle) whose net transformation is still identity, testing endpoint
closure without symmetric replay.
\textbf{T3 (Repeated-Cycle)} repeats the same closed cycle $K$ times and
tracks both endpoint-error growth and phase-aligned drift across
repetitions.
\textbf{T4 (Cascaded-Cycle)} composes two structurally different closed
cycles $C^{(1)}\circ C^{(2)}$ and asks whether residuals from the first
cycle contaminate the second.

\noindent\textbf{Scenario Settings.} Every task is evaluated under four settings along two axes---rollout
horizon and action complexity: \textbf{short-term} (125 frames, basic
in-domain case), \textbf{mid-term} (253 frames, tests error amplification
under longer generation), \textbf{long-term} (381 frames, primary
stress-test for compounding error), and \textbf{composite-action} (125
frames with multi-component controls, out-of-domain w.r.t.\ pre-training).
Matched trajectory structure across the three horizons attributes
differences to autoregressive accumulation rather than task semantics.

\noindent\textbf{Metrics.} To decouple evaluation from the CoTracker-based reward used during post-training~\cite{karaev2024cotracker}, CycleBench adopts RoMa~\cite{edstedt2024roma} as an independent dense correspondence estimator. For any two frames that should coincide, the mean 2D displacement of valid RoMa pixel matches, measured in pixels, defines the frame-pair distance
$\droma(\hat{I}_i,\hat{I}_j)$; pairs with too few matches are assigned the image diagonal. Three complementary metrics are reported:
\textbf{ESC$\downarrow$} (Endpoint State Closure), the distance between the generated endpoint and the initial frame;
\textbf{RPS$\downarrow$} (Reverse-Path Symmetry), the average distance between mirrored forward-reverse pairs;
and \textbf{RCS$\downarrow$} (Repeated-Cycle Stability), the average phase-aligned distance across cycles.

\begin{figure*}[htbp]
\centering
\includegraphics[width=1.0\linewidth]{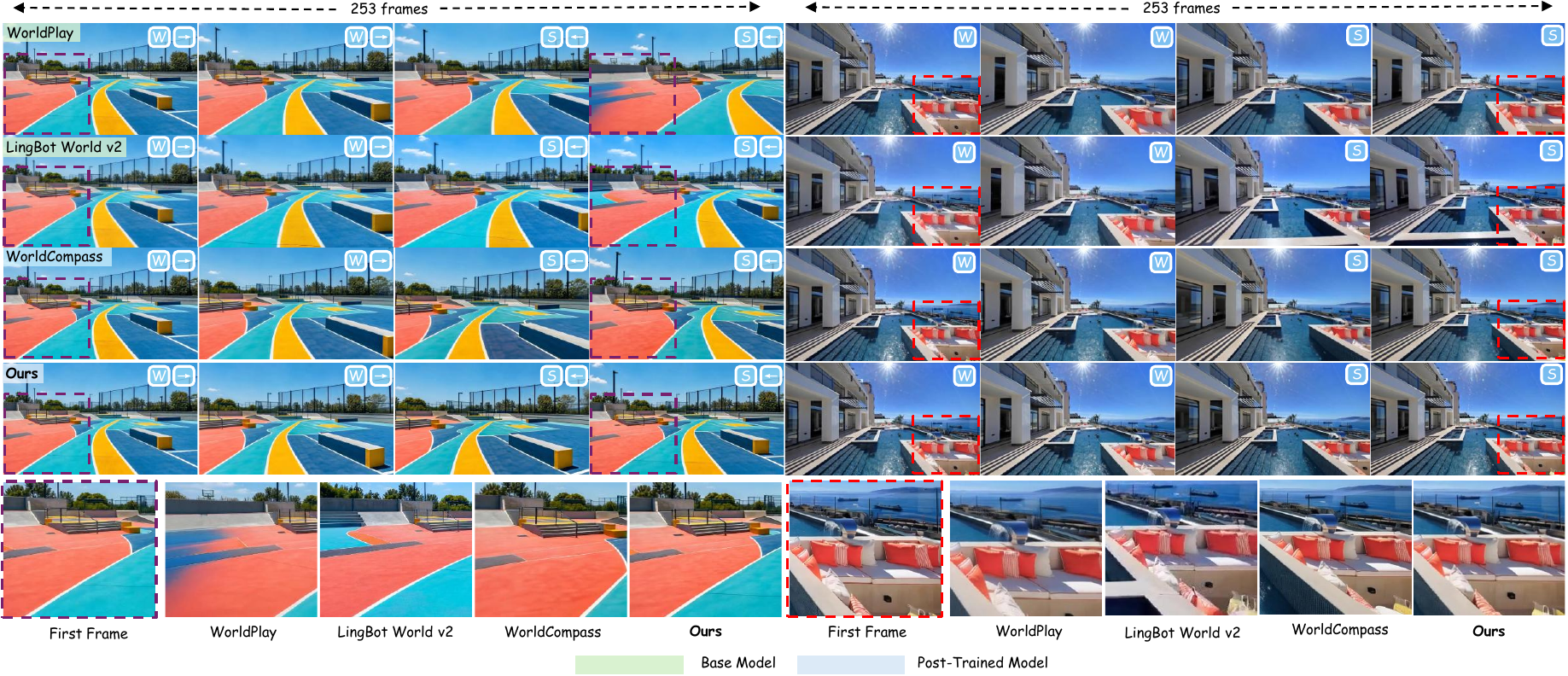}
\vspace{-20pt}
\caption{
  \textbf{Qualitative comparison on CycleBench (253 frames).}
  \emph{Left:} composite-action setting (forward$+$turn then inverse).
  \emph{Right:} simple-action setting (forward then backward).
  Each method row contains eight sampled frames in total, with four keyframes
from each setting: two from the outward phase and two from the return phase.
  The bottom strip shows a cropped zoom of the return frame alongside the
  first-frame reference for each method.
}
\vspace{-12pt}
\label{fig:qualitative}
\end{figure*}

\begin{table*}[!ht]
\centering
\footnotesize
\caption{
  \textbf{Ablation on CycleBench.} Left: basic actions across three
  horizons (short / mid / long). Right: OOD composite actions
  reported with the full metric set. \emph{Reward} isolates each cycle
  signal; \emph{Schedule} isolates the warm-up-and-combine training order.
}
\vspace{-10pt}
\label{tab:ablation}
\setlength{\tabcolsep}{8pt}
\renewcommand{\arraystretch}{1}
\begin{tabular}{l cc cc cc I ccccc}
\toprule
& \multicolumn{2}{c}{\textbf{Short-term}}
& \multicolumn{2}{c}{\textbf{Mid-term}}
& \multicolumn{2}{c I}{\textbf{Long-term}}
& \multicolumn{5}{c}{\textbf{Composite-action (OOD)}} \\
\cmidrule(lr){2-3}\cmidrule(lr){4-5}\cmidrule(lr){6-7}\cmidrule(lr){8-12}
\textbf{Variant}
  & ESC$\downarrow$ & Acc$\uparrow$
  & ESC$\downarrow$ & Acc$\uparrow$
  & ESC$\downarrow$ & Acc$\uparrow$
  & ESC$\downarrow$ & RPS$\downarrow$ & RCS$\downarrow$ & Acc$\uparrow$ & Qual$\uparrow$ \\
\midrule
\rowcolor{gray!8}Base model
  & 0.076 & 0.635 & 0.092 & 0.695 & 0.302 & 0.073
  & 0.119 & 0.088 & 0.057 & 0.136 & \phantom{0}8.98 \\
\midrule
\multicolumn{12}{l}{\textbf{Reward ablation}} \\
\rowcolor{gray!8}\quad Spatial only
  & 0.033 & 0.712 & 0.050 & 0.744 & 0.212 & 0.081
  & 0.071 & 0.054 & 0.038 & 0.473 & \phantom{0}9.77 \\
\quad Temporal only
  & 0.059 & 0.651 & 0.078 & 0.712 & 0.268 & 0.076
  & 0.098 & 0.076 & 0.048 & 0.378 & \phantom{0}9.20 \\
\midrule
\multicolumn{12}{l}{\textbf{Schedule ablation}} \\
\rowcolor{gray!8}\quad Sequential (S$\to$T)
  & 0.028 & 0.797 & 0.041 & 0.845 & 0.176 & 0.088
  & 0.061 & 0.045 & 0.030 & 0.529 & 10.21 \\
\quad Joint from scratch
  & 0.030 & 0.769 & 0.044 & 0.821 & 0.180 & 0.085
  & 0.065 & 0.052 & 0.036 & 0.510 & 10.04 \\
\midrule
\rowcolor[HTML]{D7F6FF}\textbf{Full (Ours)}
  & \textbf{0.026} & \textbf{0.833}
  & \textbf{0.038} & \textbf{0.878}
  & \textbf{0.163} & \textbf{0.095}
  & \textbf{0.057} & \textbf{0.042} & \textbf{0.026} & \textbf{0.553} & \textbf{10.42} \\
\bottomrule
\end{tabular}
\vspace{-18pt}
\end{table*}

\vspace{-8pt}
\section{Experiments}

\subsection{Implementation Details}

\noindent\textbf{Baselines.}
We compare WorldCycle against three publicly available interactive video
world models. \textbf{Lingbot World v2}~\cite{gao2026lingbotworld2} is a 14B-parameter open-source
foundation model with the largest capacity among the compared systems.
\textbf{WorldPlay} (8B, ICML~2026)~\cite{sun2025worldplay} and \textbf{WorldCompass} (ICML~2026)~\cite{wang2026worldcompass} are
the strongest smaller-scale baselines for action-conditioned generation;
WorldCompass is the closest prior work, applying reinforcement-learning
post-training to WorldPlay using per-clip inverse-pose and visual-quality
rewards. Both WorldCompass and our WorldCycle are initialized from
WorldPlay, isolating the effect of the post-training objective.

\noindent\textbf{Evaluation.}
All methods are evaluated on CycleBench (380 videos) under four task types (T1--T4) and four scenario settings: short-term (125 frames), mid-term (253 frames), long-term (381 frames), and composite-action (125 frames). Primary metrics are ESC$\downarrow$, RPS$\downarrow$, RCS$\downarrow$, and Action Accuracy$\uparrow$; HPSv3$\uparrow$ quality is reported to verify that cycle training does not degrade visual fidelity.

\noindent\textbf{Training Data and Setup.}
We post-train the 8B WorldPlay AR checkpoint on approximately 4{,}000 image--caption
pairs from real-world scenes. Each training instance samples a base or combined action
sequence and constructs closed reversible cycles and repeated executions on
the fly, so no cycle-specific supervision is annotated. Rollouts are chunked
into 16-frame video clips as in the base model. We use rollout group size
$G=16$, up to $K=8$ repeated cycles per hierarchical level, and sample one
hierarchy level per training step to keep the per-step rollout cost fixed. To optimize
the model for long-horizon generation, we set the maximum generation length to 16 clips.
The spatial warm-up runs for $t_{\mathrm{w}}=300$ steps, after which the
temporal reward is activated. The reward weights are set to
$\lambda_s=1.0$, $\lambda_t=0.5$, $\lambda_v=1.0$ and $\lambda_a=2.0$. Optimization uses AdamW with
learning rate $1\!\times\!10^{-5}$.
The training process runs 3 days on 8 H200 GPUs.

\vspace{-5pt}
\subsection{Qualitative and Quantitative Results}

Table~\ref{tab:main_results} reports results across all CycleBench tasks and settings. We break down the key findings into three distinct aspects of model performance:

\textbf{Scaling Alone Does Not Solve State Consistency}: Lingbot World v2 (14B) achieves the highest HPSv3 quality due to its larger capacity, but lags every RL post-trained baseline on cycle-consistency metrics (ESC, RPS, RCS are 1.4--8.3$\times$ worse than WorldCycle), confirming that state-returning consistency is not solved by scaling alone.

\textbf{Superiority Across Extended Horizons}: WorldCycle consistently outperforms both WorldPlay and WorldCompass. On short-term simple actions it reduces ESC by \textbf{32\%} and RPS by \textbf{44\%} over WorldCompass while maintaining comparable action accuracy (0.833 vs.\ 0.829), confirming that cycle-consistency gains do not come at the cost of local interaction following. The gains persist as the horizon extends: at 253 frames ESC drops by \textbf{21\%} and RPS by \textbf{28\%}; at 381 frames RCS falls by \textbf{34\%}, the largest relative gain across horizon  settings, reflecting the direct contribution of the temporal consistency reward to long-range stability.

\textbf{Strong Generalization to Out-of-Domain Composite Actions}: On composite actions, where the base WorldPlay's accuracy collapses to 0.136, WorldCycle achieves an accuracy of 0.553 (a \textbf{4$\times$} improvement over the base and \textbf{+11\%} over WorldCompass), confirming that cycle supervision generalizes effectively to unseen action compositions.

Figure~\ref{fig:qualitative} provides qualitative confirmation. While baselines generate locally plausible frames, they accumulate visible drift upon returning to the initial state: colors shift, geometric layouts distort, and structural details diverge. This degradation is particularly severe in the out-of-domain composite-action setting (left). WorldCycle recovers states substantially closer to the starting view in both settings, as the bottom crop strips make this visually unambiguous.

\vspace{-8pt}
\subsection{Ablation study}
\vspace{-3pt}
We analyze component contributions in Table~\ref{tab:ablation}.

\noindent\textbf{Each reward addresses a distinct failure mode.}
Removing the temporal signal primarily degrades long-range stability and composite-action generalization (e.g., long-term ESC increases from 0.163 to 0.212). Conversely, removing the spatial signal causes broader failure across all settings, because temporal supervision relies on individual cycles approximately closing; without the spatial anchor, phase-aligned comparisons become meaningless.

\noindent\textbf{Warm-up-and-combine is essential.}
The schedule ablation confirms that the ordering of the two rewards is non-trivial.
Joint-from-scratch training consistently underperforms our warm-up schedule: the temporal reward has no meaningful reference before individual cycles are grounded, so its gradient adds noise rather than signal.
Sequential training (spatial-only then temporal-only) recovers some performance but ultimately drifts because it discards the spatial constraint during the temporal phase.
Our warm-up-and-combine strategy retains the spatial anchor throughout optimization, yielding the best results across all settings.

\noindent\textbf{Training does not harm visual quality.}
Table~\ref{tab:ablation} shows that the full WorldCycle model achieves the highest HPSv3 score (10.42) among all variants,
a \textbf{16\%} gain over the base model (8.98), ruling out the hypothesis that cycle-consistency improvements are
purchased at the cost of visual fidelity.

\vspace{-5pt}
\section{Conclusion}
We presented \textbf{WorldCycle}, a self-verifiable RL framework that turns reversible action cycles into dense, annotation-free supervision for long-horizon video world models.
Spatial closure and temporal state consistency rewards jointly drive the model to learn actions as consistent state operators, overcoming reward sparsity and temporal drift while generalizing to composite action cycles outside the base model's pretraining distribution.
We further introduced \textbf{CycleBench}, the first benchmark evaluating video world models as state-transition simulators.
Extending the cycle-return principle to approximate reversibility and other action modalities such as robotic manipulation are natural directions for future work.

\clearpage
\bibliography{aaai2027}

\clearpage

\appendix
\section{Benchmark Details}

\subsection{Benchmark Overview}

CycleBench measures \textbf{how transition errors accumulate over extended
autoregressive rollouts}. For an arbitrary open action trajectory, the correct
future state is unavailable, so long-horizon error cannot be scored directly.
CycleBench sidesteps this verification bottleneck by constructing reversible
and closed action trajectories whose composed transformation is analytically
the identity. Any discrepancy between rollout states that should coincide
therefore provides an observable estimate of accumulated transition error.

The benchmark consists of 47 action trajectories evaluated from 380 initial
frames, organized as four benchmark tasks and four scenario settings. The four
tasks probe where transition error emerges, how much residual remains at
closure, how it grows across repeated executions, and whether it propagates
from one closed cycle to the next. The four settings vary rollout horizon and
action complexity, covering short-term, mid-term, long-term, and
composite-action conditions.

\noindent\textbf{Initial-Frame Collection.}
CycleBench contains 380 initial frames from two complementary sources.
We select 236 images from 4KLSDB~\cite{zhu20264klsdb}, which predominantly
cover realistic, photographic scenes. To increase the diversity of scene
content and visual styles, we additionally synthesize 144 images using
GPT Image 2. Together, the collected and generated images provide a balanced
set of realistic and synthetic environments for evaluating world-model
rollouts.

\begin{algorithm*}[t]
\caption{WorldCycle Training Process}
\label{alg:worldcycle}
\SetAlgoLined
\SetNoFillComment
\SetArgSty{textnormal}
\small{\KwIn{
World model $\pi_\theta$, EMA reference $\pi_{\theta_{\mathrm{old}}}$,
training data $\mathcal{D}$,
level weights $\boldsymbol\omega$, group size $G$, repeat count $K$,
warm-up threshold $t_{\mathrm{w}}$,
weights $\lambda_s,\lambda_t,\lambda_a,\lambda_v$
}}
\small{\KwOut{
Optimized world model $\pi_\theta$
}}

\BlankLine

\For{training step $t=1,2,\ldots$}{

{\footnotesize{\color{DarkBlue}{\tcc{Multi-scale cycle sampling: one hierarchy level per step}}}}

$k\sim\boldsymbol\omega$

\BlankLine

\For{$(o_0,a_{1:T})\in\mathcal{D}$}{

{\footnotesize{\color{DarkBlue}{\tcc{Construct closed cycle at hierarchy level $k$}}}}

$\gamma_c\leftarrow(B_k\,\|\,B_k^{-1})$;\quad
$\Gamma_K\leftarrow(\gamma_c,\dots,\gamma_c)$
{\footnotesize{\color{DarkBlue}{\tcp*{$K$ repetitions}}}}

\BlankLine

{\footnotesize{\color{DarkBlue}{\tcc{Group rollout from EMA reference ($G$ independent samples)}}}}

$\{\tau^{(i)}\}_{i=1}^{G}\sim\pi_{\theta_{\mathrm{old}}}(o_0,\Gamma_K)$

\BlankLine

{\footnotesize{\color{DarkBlue}{\tcc{Compute per-rollout rewards}}}}

\For{$i=1,\ldots,G$}{

$R_s^{(i)}\leftarrow\mathrm{SpatialClosure}(\tau^{(i)})$
{\footnotesize{\color{DarkBlue}{\tcp*{Dense mirrored-pair verification}}}}

$R_t^{(i)}\leftarrow\mathbf{1}[t\!\geq\!t_{\mathrm{w}}]\cdot\mathrm{TemporalStationarity}(\tau^{(i)})$
{\footnotesize{\color{DarkBlue}{\tcp*{Activated after spatial warm-up}}}}

$R^{(i)}\leftarrow\lambda_s R_s^{(i)}+\lambda_t R_t^{(i)}+\lambda_a R_{\mathrm{act}}^{(i)}+\lambda_v R_{\mathrm{HPS}}^{(i)}$

}

\BlankLine

{\footnotesize{\color{DarkBlue}{\tcc{Group-normalize rewards and optimize (DiffusionNFT)}}}}

$r^{(i)}\leftarrow\operatorname{Norm}\!\bigl(\{R^{(i)}\}_{i=1}^{G}\bigr)$
{\footnotesize{\color{DarkBlue}{\tcp*{$(R^{(i)}-\mu)/\sigma$ within group}}}}

$\mathcal{L}_{\mathrm{RL}}\leftarrow
\dfrac{1}{G}\displaystyle\sum_{i=1}^{G}\!\Bigl[
r^{(i)}\bigl\lVert v_\theta^{+}\!-\!v^{(i)}\bigr\rVert_2^2
+(1\!-\!r^{(i)})\bigl\lVert v_\theta^{-}\!-\!v^{(i)}\bigr\rVert_2^2
\Bigr]$

\BlankLine

$\theta\leftarrow\mathrm{AdamW}(\theta,\nabla_\theta\mathcal{L}_{\mathrm{RL}})$;\quad
$\theta_{\mathrm{old}}\leftarrow\mathrm{EMA}(\theta_{\mathrm{old}},\theta)$
{\footnotesize{\color{DarkBlue}{\tcp*{Update model and reference}}}}

}

}

\end{algorithm*}

\subsection{Benchmark Tasks}

Each task uses analytically known frame correspondences within closed
trajectories to probe a distinct aspect of accumulated transition error.

\paragraph{T1: Reversible-Cycle.}
T1 executes a forward action sequence
\(H=(a_1,\ldots,a_K)\), followed by its exact inverse
\(H^{-1}=(a_K^{-1},\ldots,a_1^{-1})\). Under ideal dynamics, the complete
forward--inverse trajectory induces the identity transformation. Moreover,
each state reached during the forward segment should coincide with its
mirrored state during the inverse segment. T1 therefore evaluates whether
mirrored forward--reverse frames coincide, providing a dense path-wise probe
that localizes where transition error first appears and how it develops along
the rollout.

\paragraph{T2: Closed-Cycle.}
T2 executes a non-retracing closed action sequence \(C\), such as a rectangular
or polygonal path. Although the trajectory does not symmetrically replay its
earlier path, its ideal net transformation is still the identity, so the
generated endpoint should coincide with the initial observation. T2 therefore
tests endpoint closure after a structurally closed but non-symmetric
trajectory, preventing success through simple reverse-path replay.

\paragraph{T3: Repeated-Cycle.}
T3 repeats the same closed cycle \(C\) for \(K\) executions. If one cycle has
length \(L\), the endpoints at times \(L,2L,\ldots,KL\) should all coincide
with the initial observation, while phase-aligned states across executions
should remain consistent. T3 tracks both endpoint-error growth and
phase-aligned drift, directly measuring whether small residuals are amplified
as the rollout horizon increases.

\paragraph{T4: Cascaded-Cycle.}
T4 sequentially executes two structurally different closed cycles,
\(C^{(1)}\) and \(C^{(2)}\), where the second begins from the generated
endpoint of the first:
\[
I_0 \xrightarrow{C^{(1)}} \hat{I}_{T_1}
\xrightarrow{C^{(2)}} \hat{I}_{T_1+T_2}.
\]
Because both cycles ideally induce identity transformations, the intermediate
and final endpoints should each coincide with \(I_0\). T4 asks whether the
residual produced by the first cycle contaminates the second, thereby probing
cross-cycle error propagation under different action compositions.

\subsection{Scenario Settings}

Every task is evaluated under four settings along two complementary axes:
rollout horizon and action complexity. The short-, mid-, and long-term
settings preserve matched trajectory structures while progressively extending
the rollout, so differences can be attributed primarily to autoregressive
error accumulation rather than to different task semantics. The
composite-action setting instead tests more complex, multi-component controls.

\paragraph{Short-term.}
The short-term setting contains 125-frame rollouts and serves as the basic
in-domain case for measuring where transition error first emerges and how much
residual remains after a relatively short reversible or closed trajectory.

\paragraph{Mid-term.}
The mid-term setting extends the rollout to 253 frames while preserving the
corresponding trajectory structure. It tests whether residuals observed at
short horizons are amplified under longer autoregressive generation.

\paragraph{Long-term.}
The long-term setting further extends the rollout to 381 frames and serves as
the primary stress test for compounding error, including endpoint residuals,
reverse-path discrepancies, and repeated-cycle drift.

\paragraph{Composite-action.}
The composite-action setting uses 125-frame rollouts with controls that combine
multiple motion components, such as simultaneous translation and rotation.
These controls are out of domain with respect to the base model's pretraining
distribution and test whether cycle consistency generalizes to unseen action
compositions rather than only to primitive controls.

\subsection{RoMa Correspondence Distance}

To decouple evaluation from the CoTracker-based correspondence reward used
during post-training, CycleBench adopts RoMa as an independent dense
correspondence estimator. Given two generated frames \(\hat{I}_i\) and
\(\hat{I}_j\), RoMa produces correspondences
\(\{(\mathbf{p}^{(i)}_k,\mathbf{p}^{(j)}_k,c_k)\}_{k=1}^{N_{ij}}\), where
\(\mathbf{p}^{(i)}_k,\mathbf{p}^{(j)}_k\in\mathbb{R}^2\) are matched pixel
coordinates and \(c_k\) is the associated confidence. We retain the valid
matches
\(\mathcal{V}_{ij}=\{k\mid c_k\geq\tau_{\mathrm{match}}\}\) and define the
frame-pair distance as
\begin{equation}
\droma(\hat{I}_i,\hat{I}_j)
=
\frac{1}{|\mathcal{V}_{ij}|}
\sum_{k\in\mathcal{V}_{ij}}
\left\|
\mathbf{p}^{(i)}_k-\mathbf{p}^{(j)}_k
\right\|_2 .
\end{equation}
For frame pairs that should correspond to the same underlying state, this
distance provides an observable estimate of the accumulated transition
residual. A smaller value indicates better alignment between the matched
visual structures.

Because all evaluated videos share the same resolution, CycleBench reports
the displacement directly in pixels. To prevent failed matching from being
interpreted as low error, frame pairs with fewer than \(N_{\min}\) valid
correspondences are assigned the maximum image-space distance
\(D_{\mathrm{img}}=\sqrt{H^2+W^2}\), where \(H\) and \(W\) denote the frame
height and width.

\subsection{Evaluation Metrics}

CycleBench reports three complementary correspondence-based metrics, ordered
as in the main results table: \ESC{} measures the final closure residual,
\RPS{} measures discrepancies along mirrored forward--reverse paths, and
\RCS{} measures phase-aligned drift across cycle executions. Lower values
indicate better state-returning consistency.

\paragraph{Endpoint State Closure (\ESC).}
\ESC{} is the distance between the generated endpoint and the initial frame:
\begin{equation}
\mathcal{L}_{\ESC}^{(\xi)}
=
\droma(I_0,\hat{I}_T).
\end{equation}
It summarizes the total residual remaining after the complete closed
trajectory as a single global closure error.

\paragraph{Reverse-Path Symmetry (\RPS).}
\RPS{} is the average distance between mirrored forward--reverse frame pairs.
For a standard T1 trajectory \(\xi\) of length \(T=2K\), let
\[
\mathcal{P}^{\mathrm{sym}}_{\xi}
=
\{(t,T-t)\mid 1\leq t<K\}
\]
denote the set of non-trivial mirrored pairs. We define
\begin{equation}
\mathcal{L}_{\RPS}^{(\xi)}
=
\frac{1}{|\mathcal{P}^{\mathrm{sym}}_{\xi}|}
\sum_{(i,j)\in\mathcal{P}^{\mathrm{sym}}_{\xi}}
\droma(\hat{I}_i,\hat{I}_j).
\end{equation}
Unlike an endpoint-only measure, \RPS{} provides a dense path-wise diagnostic
of where residual error emerges during reversible generation.

\paragraph{Repeated-Cycle Stability (\RCS).}
\RCS{} is the average distance between phase-aligned states across cycle
executions. For T3, we anchor all later executions to the first cycle. If each
cycle has length \(L\), the correspondence set is
\[
\mathcal{P}^{\mathrm{cyc}}_{\xi}
=
\{(\ell,(r-1)L+\ell)\mid r=2,\ldots,K,\;
\ell=0,\ldots,L\}.
\]
The metric is
\begin{equation}
\mathcal{L}_{\RCS}^{(\xi)}
=
\frac{1}{|\mathcal{P}^{\mathrm{cyc}}_{\xi}|}
\sum_{(i,j)\in\mathcal{P}^{\mathrm{cyc}}_{\xi}}
\droma(\hat{I}_i,\hat{I}_j).
\end{equation}
For T4, the same principle is applied to analytically coincident closure
states at the cycle handoff and after the second cycle. A low \RCS{} indicates
that repeated or cascaded cycle execution does not progressively amplify
transition residuals.

\paragraph{Auxiliary Metrics.}
In addition to the three state-consistency metrics, CycleBench reports Action
Accuracy, using the action-following evaluator adopted by WorldCompass, and
HPSv3 visual quality. These auxiliary metrics verify that improvements in
cycle consistency do not come at the cost of local action following or visual
fidelity.

\begin{table*}[t]
\centering
\footnotesize
\caption{
  \textbf{VBench video quality comparison across four settings.} All six dimensions are higher-is-better ($\uparrow$).
  Results are weighted averages over T1--T4 tasks within each setting (380 videos total).
  WorldCycle consistently matches or exceeds WorldCompass and improves over WorldPlay across all settings.
}
\vspace{-8pt}
\label{tab:vbench}
\resizebox{\linewidth}{!}{%
\renewcommand{\arraystretch}{1.25}
\setlength{\minrowclearance}{2pt}
\setlength{\tabcolsep}{4pt}
\begin{tabular}{l cccccc I cccccc}
\toprule
& \multicolumn{6}{c I}{\textbf{Short-term (125 frames)}}
& \multicolumn{6}{c}{\textbf{Mid-term (253 frames)}} \\
\cmidrule(lr){2-7}\cmidrule(lr){8-13}
\rowcolor{mygray}\textbf{Method}
  & Aesth.$\uparrow$ & S.Cons$\uparrow$ & B.Cons$\uparrow$ & Img$\uparrow$ & T.Flick$\uparrow$ & Mot$\uparrow$
  & Aesth.$\uparrow$ & S.Cons$\uparrow$ & B.Cons$\uparrow$ & Img$\uparrow$ & T.Flick$\uparrow$ & Mot$\uparrow$ \\
\midrule
\rowcolor{gray!8}WorldPlay \textit{\scriptsize(8B, ICML'26)}
  & 0.653 & 0.950 & 0.948 & 0.744 & \textbf{0.956} & 0.985
  & 0.631 & 0.930 & 0.939 & 0.723 & \textbf{0.955} & 0.983 \\
WorldCompass \textit{\scriptsize(ICML'26)}
  & 0.676 & 0.963 & 0.953 & 0.770 & 0.946 & 0.983
  & 0.665 & 0.954 & 0.948 & \textbf{0.772} & 0.944 & 0.983 \\
\hdashline
\rowcolor[HTML]{D7F6FF}\textbf{WorldCycle}
  & \textbf{0.676} & \textbf{0.968} & \textbf{0.955} & \textbf{0.773} & 0.945 & \textbf{0.985}
  & \textbf{0.665} & \textbf{0.961} & \textbf{0.948} & 0.771 & 0.943 & \textbf{0.984} \\
\specialrule{1.2pt}{2pt}{2pt}
& \multicolumn{6}{c I}{\textbf{Composite-action (125 frames)}}
& \multicolumn{6}{c}{\textbf{Long-term (381 frames)}} \\
\cmidrule(lr){2-7}\cmidrule(lr){8-13}
\rowcolor{mygray}\textbf{Method}
  & Aesth.$\uparrow$ & S.Cons$\uparrow$ & B.Cons$\uparrow$ & Img$\uparrow$ & T.Flick$\uparrow$ & Mot$\uparrow$
  & Aesth.$\uparrow$ & S.Cons$\uparrow$ & B.Cons$\uparrow$ & Img$\uparrow$ & T.Flick$\uparrow$ & Mot$\uparrow$ \\
\midrule
\rowcolor{gray!8}WorldPlay \textit{\scriptsize(8B, ICML'26)}
  & 0.637 & 0.948 & 0.943 & 0.733 & \textbf{0.960} & \textbf{0.985}
  & 0.579 & 0.818 & 0.891 & 0.572 & \textbf{0.980} & \textbf{0.989} \\
WorldCompass \textit{\scriptsize(ICML'26)}
  & 0.666 & 0.960 & 0.948 & 0.762 & 0.943 & 0.981
  & 0.665 & 0.871 & 0.921 & 0.707 & 0.958 & 0.984 \\
\hdashline
\rowcolor[HTML]{D7F6FF}\textbf{WorldCycle}
  & \textbf{0.672} & \textbf{0.966} & \textbf{0.948} & \textbf{0.768} & 0.944 & 0.984
  & \textbf{0.697} & \textbf{0.907} & \textbf{0.925} & \textbf{0.729} & 0.961 & 0.988 \\
\bottomrule
\end{tabular}%
}
\vspace{-10pt}
\end{table*}

\section{WorldCycle Training Algorithm}

Algorithm~\ref{alg:worldcycle} summarizes the complete training procedure of
WorldCycle. At each training step, we sample a hierarchy level and construct
a closed action cycle by concatenating an action block with its inverse,
which is then repeated to form a long-horizon trajectory. The EMA reference
model generates a group of independent rollouts, from which we compute the
spatial closure, temporal consistency, action-following, and visual-quality
rewards. The temporal reward is activated only after the spatial warm-up
stage, while both cycle rewards remain jointly active thereafter. The
group-normalized rewards are finally used in a DiffusionNFT-style objective
to update the world model, followed by an EMA update of the reference model.

\begin{figure}[t]
\centering
\includegraphics[width=\linewidth]{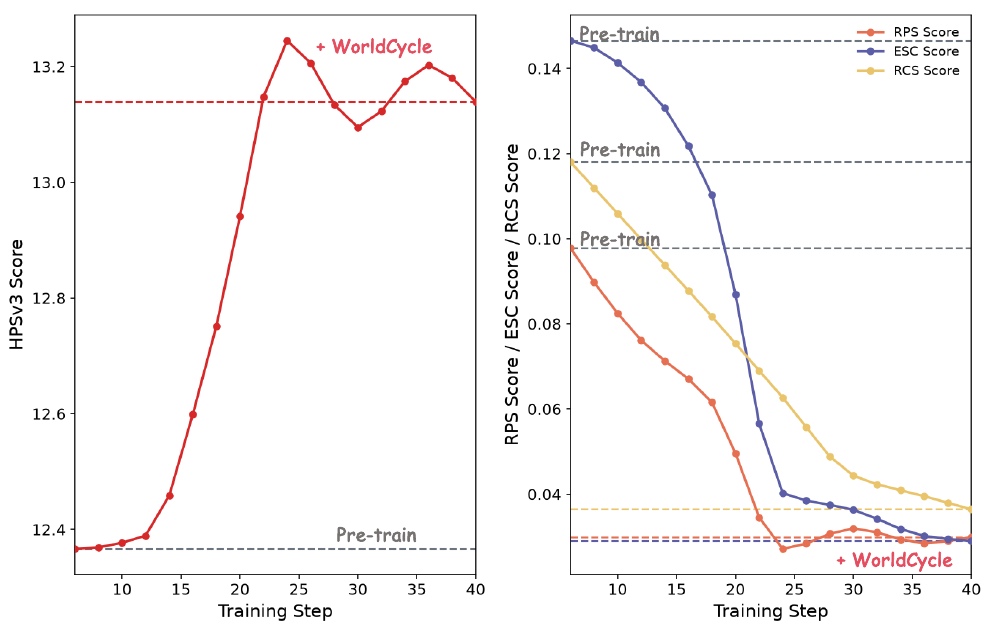}
\caption{
  Evolution of cycle consistency and visual quality metrics during the RL
  training of WorldCycle. These metrics are evaluated on a fixed CycleBench
  validation subset at every training step. 
}
\label{fig:training_curve}
\end{figure}

\section{Additional Analysis Details}

\noindent\textbf{Training Dynamics.}
Figure~\ref{fig:training_curve} further examines the optimization process
when WorldCycle is initialized from WorldPlay. All three cycle-consistency
metrics decrease steadily during post-training and converge below their
initial values. Meanwhile, HPSv3 increases and remains above the WorldPlay
initialization after convergence. These trajectories indicate that the
combined objective provides stable optimization of long-horizon consistency
without sacrificing visual quality.

\begin{figure*}[t]
    \centering
    \includegraphics[width=\linewidth]{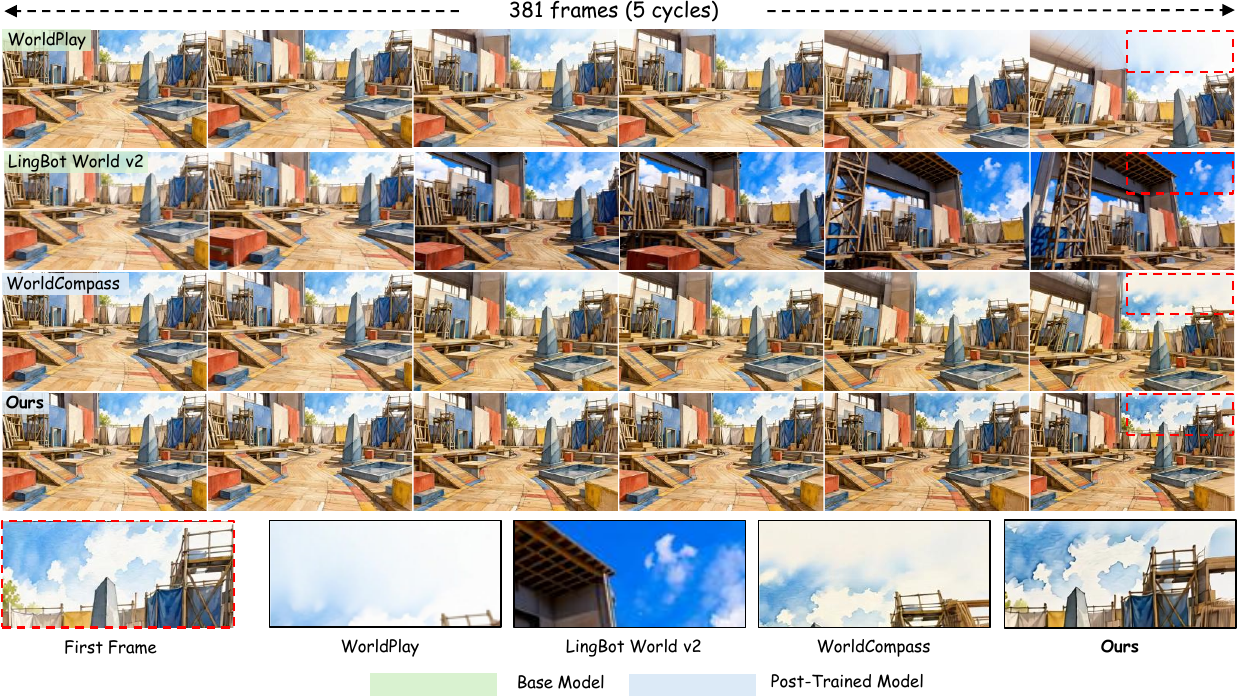}
    \caption{
    \textbf{Long-horizon qualitative comparison over five repeated cycles
    (381 frames).}
    Each row shows sampled frames from the same long-term rollout. The bottom
    row enlarges the red-boxed region in the final return frame and compares it
    with the corresponding region in the first frame. LingBot World v2 exhibits
    severe color drift, most visibly through the increasingly saturated blue
    sky, whereas WorldPlay and WorldCompass develop pronounced visual artifacts,
including whitening in the sky and unnatural linear patterns around
the rooftop structure. 
In contrast, WorldCycle preserves stable colors, textures,
    and scene content, yielding a final return frame that remains close to the
    initial observation after all five cycles.
    }
    \label{fig:long_term_qualitative}
\end{figure*}

\noindent\textbf{Long-Horizon Qualitative Stability.}
Figure~\ref{fig:long_term_qualitative} provides a qualitative analysis under
the 381-frame long-term setting, in which the model executes five repeated
cycles. The baselines exhibit distinct forms of accumulated visual error.
LingBot World v2 undergoes pronounced color drift, with the sky becoming
progressively oversaturated and unnaturally blue. WorldPlay and WorldCompass
instead develop pronounced visual artifacts in the sky region, including
whitening and unnatural linear patterns in the final return frame. In
contrast, WorldCycle maintains stable colors, textures, and scene content
throughout the rollout, producing a final return frame that remains closely
aligned with the initial observation. These results demonstrate that the
cycle-consistency rewards effectively suppress the accumulation of both color
drift and visual artifacts over repeated long-horizon transitions.

\section{VBench Video Quality Evaluation}

To verify that WorldCycle's training does not degrade
perceptual video quality, we evaluate all methods on six VBench
dimensions~\cite{huang2024vbench}: Aesthetic Quality, Subject Consistency,
Background Consistency, Imaging Quality, Temporal Flickering, and Motion
Smoothness. These dimensions are computed independently of the CycleBench
reward, providing a complementary quality signal. Lingbot-World-2 VBench
scores are omitted here due to limited sample size (12 videos); overall
VBench scores for Lingbot-World-2 are: Aesthetic 0.743, Subj.\ Cons.\
0.882, Bg.\ Cons.\ 0.923, Imaging 0.746, Temp.\ Flick.\ 0.961, Motion
0.979.

WorldCycle matches or exceeds WorldCompass on nearly all dimensions across
all four settings, and consistently outperforms the base WorldPlay model.
Subject Consistency and Imaging Quality show the largest gains over WorldPlay
across all settings, consistent with the observation that reducing transition
residuals also stabilizes frame-level appearance. The improvements are
sustained in the composite-action setting, confirming that cycle training on
out-of-distribution action sequences does not introduce quality degradation.

\section{Discussion and Limitations}

WorldCycle does not treat reversible action cycles merely as a specialized
state-returning task. Instead, it uses them as a controlled mechanism for
exposing and correcting long-horizon transition errors that would otherwise
be difficult to observe. Because the composed transformation of a closed
trajectory is analytically known, mirrored and repeated states provide dense
verification without requiring ground-truth video supervision. The consistent
improvements across different rollout horizons and action structures indicate
that reducing local transition residuals improves the overall trajectory-level
dynamics. Moreover, the gains on out-of-distribution composite actions suggest
that the learned dynamics increasingly capture the underlying compositional
structure of actions rather than simply memorizing specific action programs.
Closed cycles should therefore be understood as a self-verifiable supervision
mechanism, rather than a restriction on the model's applicable trajectory
space.

The main limitation of WorldCycle is that its exact supervision signals rely
on action structures with explicit inverse operations or analytically known
closure relations. For irreversible or partially irreversible processes, such
as object deformation, contact-rich interaction, and persistent state changes,
an action sequence cannot generally recover its initial state by simply
applying inverse controls. The current cycle construction therefore cannot be
directly applied to these dynamics. Extending self-verifiable supervision to
irreversible processes through conservation laws, equivalent terminal states,
or other analytically verifiable constraints represents an important direction
for future work.

\section{Relation to Concurrent Cycle-Based Methods}

Two concurrent works explore cyclic constraints for improving video world
models from complementary perspectives. Cycle-World~\cite{su2026cycleworld}
introduces reverse-prediction cycle consistency for long-horizon video
generation: an auxiliary reverse model reconstructs the preceding latent
chunk from the current generated chunk and is further reused for
inference-time latent correction. World Models as Group
Actions~\cite{wang2026worldmodels} instead formalizes action-conditioned
dynamics through group actions and enforces identity, inverse, and
composition consistency using synthesized latent-space supervision.

Despite these high-level similarities, the central distinction lies in
\textbf{the role of cyclic structure in the optimization pipeline}. Both
concurrent approaches formulate cycle consistency as a
\textbf{differentiable loss for model training or fine-tuning}. In contrast,
WorldCycle focuses specifically on
\textbf{reinforcement-learning post-training of a pretrained interactive
world model}. Our action cycles are not introduced merely as another
structural regularizer. Their analytically known closure relations instead
serve as \textbf{self-verifiable reward functions}, converting sampled
long-horizon rollouts into annotation-free scalar feedback. This directly
addresses the verification bottleneck in world-model post-training, where
ground-truth future videos are generally unavailable.

The three methods also differ in how their cycles are instantiated.
Cycle-World forms a temporal prediction loop between adjacent latent chunks,
whereas World Models as Group Actions regularizes the latent dynamics to
respect predefined algebraic relations. WorldCycle constructs closed action
programs in the environment and evaluates their complete generated
trajectories using spatial closure and temporal consistency rewards. Thus,
the cyclic structure serves three distinct purposes:
\textbf{reverse-prediction regularization}, \textbf{group-action
regularization}, and \textbf{self-verifiable reward construction for RL
post-training}, respectively.

\end{document}